\documentclass{article}

\PassOptionsToPackage{numbers, compress}{natbib}
\usepackage[preprint]{nips_2018}

\usepackage[utf8]{inputenc} 
\usepackage[T1]{fontenc}    
\usepackage{hyperref}       
\usepackage{url}            
\usepackage{booktabs}       
\usepackage{amsfonts}       
\usepackage{nicefrac}       
\usepackage{microtype}      

\usepackage{amssymb}
\usepackage{amsmath}
\usepackage{graphicx}
\usepackage{multirow}
\usepackage{caption}
\usepackage{subfigure}
\usepackage{bm}

\title{Geometry-Contrastive GAN for Facial Expression Transfer}

\author{
	Fengchun Qiao, 
	Naiming Yao, 
	Zirui Jiao, 
	Zhihao Li,
	Hui Chen, 
	Hongan Wang
	\\ 
	Institute of Software, Chinese Academy of Sciences \\
}

\begin{document}

\maketitle

\begin{abstract}

In this paper, we propose a Geometry-Contrastive Generative Adversarial Network (GC-GAN) for transferring continuous emotions across different subjects. Given an input face with certain emotion and a target facial expression from another subject, GC-GAN can generate an identity-preserving face with the target expression. Geometry information is introduced into cGANs as continuous conditions to guide the generation of facial expressions. In order to handle the misalignment across different subjects or emotions, contrastive learning is used to transform geometry manifold into an embedded semantic manifold of facial expressions. Therefore, the embedded geometry is injected into the latent space of GANs and control the emotion generation effectively. Experimental results demonstrate that our proposed method can be applied in facial expression transfer even there exist big differences in facial shapes and expressions between different subjects. 

\end{abstract}

\section{Introduction}
Facial expression transfer aims to transfer facial expressions from a source subject to a target subject. The newly-synthesized expressions of the target subject are supposed to be identity-preserving and exhibit similar emotions to the source subject. A wide range of applications such as virtual reality, affective interaction, and artificial agents could be facilitated by the advances of facial expression transfer.
The past few decades has witnessed numerous techniques designed for facial expression transfer. Most representative methods \cite{ Dale11_reenactment,Thies15_realtime,Thies16_reenactment,Zeiler11_rbm} are conducted in a sequence-to-sequence way where they assume the availability of a video of the target subject with variants in facial expressions and head poses, which limits their applications. Averbuch \textit{et al}. \cite{Averbuch17_portraits} proposed a warp-based method to animate a single facial image in a sequence-to-image way, capable of generating both photo-realistic and high-resolution facial expression videos, just like bringing a still portrait to life. However, there exists misalignment across different subjects without the assistance of neutral faces for alignment, leading to artifacts on the generated faces. Moreover, large manual adjustments or high computations are still required to generate realistic expressions. How to handle pixel-wise misalignments across different subjects with different emotions in an easy and automatic way is still an open problem.

Recently, Generative Adversarial Nets (GANs) \cite{Goodfellow14_GAN} have received extensive attentions to generate lavish and realistic faces. Conditional GANs (cGANs) have also been widely applied in many face-related tasks, such as face pose manipulation \cite{Yin17_FFGAN}, face aging \cite{Zhang17_age}, and facial expression transfer \cite{Choi17_StarGAN,Hui18_ExprGAN,Song17_G2GAN,Zhou17_FES}. Existing works relating to GAN-based facial expression transfer mainly focus on generating facial expressions with discrete and limited emotion states. Emotion states (\textit{e.g.}, \textit{happy}, \textit{sad}) were usually encoded as conditions of one-hot codes to control the generation of expressions \cite{Choi17_StarGAN,Hui18_ExprGAN,Zhou17_FES}. Facial landmarks of action units (AU) in the Facial Action Coding System (FACS) \cite{Ekman77_FACS} were also adopted as conditions to guide the generated expressions \cite{Song17_G2GAN}. 
In \cite{Song17_G2GAN}, the target face and facial landmarks of the driving face are concatenated in image space, which brings extra artifacts when there exist big differences in facial shapes and expressions between target and driving faces. Since people express emotions in a continuous and vivid way, how to inject facial geometry into GANs to generate continuous emotions of the target subject is essential in facial expression transfer.

In order to transfer continuous emotions across different subjects, a Geometry-Contrastive Generative Adversarial Network (GC-GAN) is proposed in this paper. GC-GAN consists of a facial geometry embedding network, an image generator network, and an image discriminator network. Contrastive learning from geometry information is integrated in embedding network. Its bottleneck layer, representing a semantic manifold of facial expressions, is concatenated into the latent space in GAN. Therefore, continuous facial expressions are displayed within the latent space, and the pixel-to-pixel misalignments across different subjects are resolved via embedded geometry. Our main contributions are as follows:
(1) We apply contrastive learning in GAN to embed geometry information onto a semantic manifold in the latent space.
(2) We inject facial geometry to guide the facial expression transfer across different subjects with Lipschitz continuity.
(3) Experimental results demonstrate that our proposed method can be applied in facial expression transfer even there exist big differences in facial shapes and expressions between different subjects.

\section{Related work}


GAN-based conditional image generation has been actively studied. Mirza \textit{et al}.\cite{Mirza14_cGAN} proposed conditional GAN (cGAN) to generate images controlled by labels or attributes. Larsen \textit{et al}.\cite{Larsen16_VAEGAN} proposed VAE-GAN combing variational autoencoder and GAN to learn an embedding representation which can be used to modify high-level abstract visual features. cGANs have also been applied in facial expression transfer \cite{Choi17_StarGAN,Hui18_ExprGAN,Song17_G2GAN,Zhou17_FES}. Zhou \textit{et al}.\cite{Zhou17_FES} proposed a conditional difference adversarial autoencoder (CDAAE) to generate faces conditioned on emotion states or AU labels. Choi \textit{et al}.\cite{Choi17_StarGAN} proposed StarGAN to perform image-to-image translations for multiple domains using only a single model, which can also be applied in facial expression transfer. Previous studies mainly focus on generating facial images conditioned on discrete emotion states. However, human emotion is expressed in a continuous way, thus discrete states are not sufficient to describe detailed characteristics of facial expressions.

Some researchers have attempted to incorporate continuous information such as geometry into cGANs \cite{Balakrishnan18_synthesizePose,Kossaifi17_GAGAN,Ma17_PG,Song17_G2GAN,Wang18_every}. Ma \textit{et al}.\cite{Ma17_PG} proposed a pose guided person generation network (PG$^2$), which allows to generate images of individuals in arbitrary poses. The target pose is defined by a set of 18 joint locations and encoded as heatmaps, which are concatenated with the input image in PG$^2$. In addition, they adopted a two-stage generation method to enhance the quality of generated images. 
Song \textit{et al}.\cite{Song17_G2GAN} proposed a Geometry-Guided Generative Adversarial Network (G2GAN), which applies facial geometry information to guide the transfer of facial expressions. In G2GAN, facial landmarks are treated as an additional image channel to be concatenated with the input face directly. They use dual generators to perform the synthesis and the removal of facial expressions simultaneously. The neutral facial images are generated through the removal network and used for the subsequent facial expression transfer. This procedure brings additional artifacts and degrades the performance especially when the driving faces are collected from other subjects in different emotions.

Contrastive learning \cite{Chopra05_contrastive} minimizes a discriminative loss function that drives the similarity metric to be small for pairs of the same class, and large for pairs from different classes. Although contrastive learning has been widely used in recognition tasks \cite{Veit17_CSN,Wang17_metric,Zhang17_fer}, potentials of the semantic manifold in latent space haven't been fully explored, which means that inter-class transitions can be expressed in a continuous representation. In our method, we introduce contrastive learning in GANs to embed geometry information and image appearance in continuous latent space to smoothly minimize the pixel-wise misalignments between different subjects and to generate continuous emotion expressions.

\section{Method}

\subsection{Overview}

In the following, we supposed that $\bm{I}_i^e$ and $\bm{g}_i^e$ represent a facial image and a geometry of facial landmarks respectively, where the superscript $e$ indicates an emotion state while the subscript $i$ denotes an identity. Given a facial image $\bm{I}^u_j$ of a subject $j$ with emotion $u$ and a geometry variable $\bm{g}^v_k$ of a subject $k$ with target emotion $v$, GC-GAN aims to generate a new identity-preserving facial image $\hat{\bm{I}}^{v}_j$ conditioned on the target facial expression. Our problem is described as:
\begin{equation}
\{\bm{I}^u_j,\bm{g}^v_k\} \longrightarrow \hat{\bm{I}}^{v}_j
\label{problem}
\end{equation}

\begin{figure}[t]
	\centering
	\includegraphics[width=12cm]{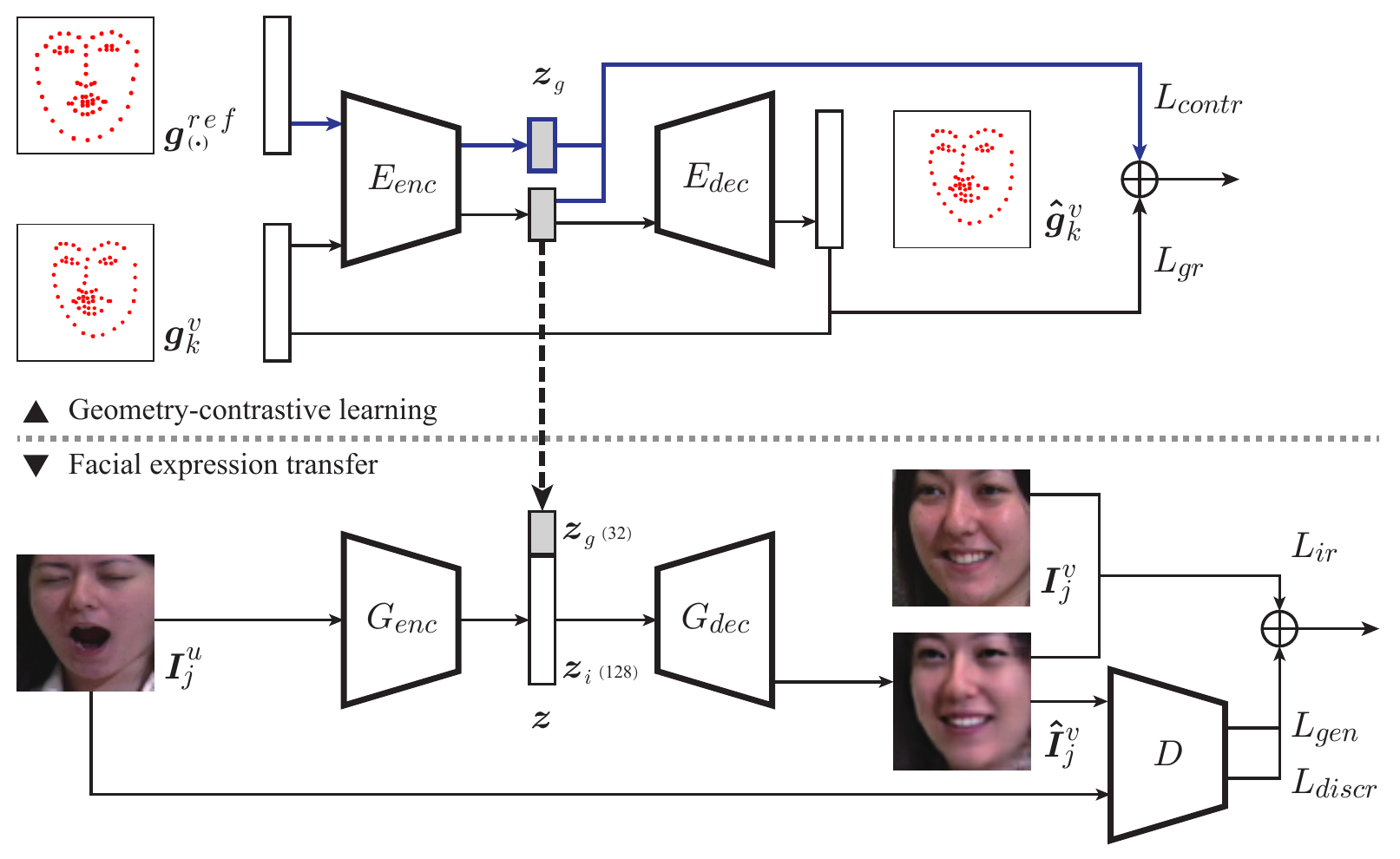}
	\caption{Overview of GC-GAN framework. GC-GAN consists of three components: facial geometry embedding network $E$ ($E_{enc}$, $E_{dec}$), image generator network $G$ ($G_{enc}$, $G_{dec}$), and image discriminator network $D$.}
	\label{fig_arch}
	\vskip-2.5mm
\end{figure}

The overall framework of GC-GAN is shown in Figure \ref{fig_arch}. GC-GAN consists of three components: a facial geometry embedding network $E$ ($E_{enc}$, $E_{dec}$), an image generator network $G$ ($G_{enc}$, $G_{dec}$), and an image discriminator network $D$. Input facial image $\bm{I}^u_j$ and target landmarks $\bm{g}^v_k$ are encoded by $G_{enc}$ and $E_{enc}$ into $\bm{z}_i$ and $\bm{z}_g$ respectively. Then $\bm{z}_i$ and $\bm{z}_g$ are concatenated into a single vector $\bm{z}$ for $G_{dec}$. $\bm{g}^{ref}_{(\cdot)}$ is reference facial landmarks used for contrastive learning against $\bm{g}^v_k$. Note that geometry expression features $\bm{z}_g$ and image identity features $\bm{z}_i$ are learned in a disentangled manner, thus we could modify expression and keep identity preserved.

\subsection{Geometry-Contrastive GAN}

The objective of original GANs \cite{Goodfellow14_GAN} is formulated as:
\begin{equation}
\label{eqgan}
L_{GAN} = 
\operatorname*{\mathbb{E}}\limits_{\bm{x} \sim p_{data}} [\log D(\bm{x})] + 
\operatorname*{\mathbb{E}}\limits_{\bm{z} \sim p_z} [\log(1-D(G(\bm{z})))],
\end{equation}
where $D$ and $G$ are the discriminator and generator respectively. In order to guide the generated samples, conditional information is input into generators as well as input prior $\bm{z}$ in GANs. Classical and other cGANs mainly use a method of concatenation of prior variables and conditional variables, which is often represented as one-hot codes \cite{Mirza14_cGAN} or embedded discrete variables \cite{Reed16_draw}.

Now we first introduce continuous conditions into generators to provide more precise guidance. The input of generator $\bm{z}$ consists of two parts, $\bm{z}_i$ and $\bm{z}_g$, \textit{i.e.}, $\bm{z} = \left( \bm{z}_i, \bm{z}_g \right)$. $\bm{z}_i$ and $\bm{z}_g$ obey different independent prior distributions $p_i$ and $p_g$, so $\bm{z}\sim p_z=p_i p_g$, and thus $\bm{z}$ is also a prior. When we use the prior to drive the generation of images, it is usually accepted that the generated  images are not abruptly changed when the input $\bm{z}$ is gradually changed. Referring to the LS-GAN \cite{Qi_LSGAN}, this assumption can be represented as the Lipschitz continuity of $G(\bm{z})$ for $\bm{z}$, \textit{i.e.}, $\forall \bm{z} = \left( \bm{z}_i, \bm{z}_g \right)$, $\bm{z}' = \left( \bm{z}_i', \bm{z}_g' \right)$, $\exists \rho > 0$, s.t. $\vert G(\bm{z}) - G(\bm{z}') \vert \le \rho \Vert \bm{z} - \bm{z}' \Vert$.
When $\bm{z}_g$ is fixed, \textit{i.e.}, $\bm{z}_g = \bm{z}_g'$, we get
$\vert G(\bm{z}_i, \bm{z}_g) - G(\bm{z}_i', \bm{z}_g) \vert \le \rho \Vert \bm{z}_i - \bm{z}_i' \Vert$,
which means $G(\bm{z}_i, \bm{z}_g)$ is Lipschitz continuous for $\bm{z}_i$.
Similarly, when $\bm{z}_i$ is fixed, it is concluded that $G(\bm{z}_i, \bm{z}_g)$ is Lipschitz continuous for $\bm{z}_g$:
\begin{equation}
\vert G(\bm{z}_i, \bm{z}_g) - G(\bm{z}_i, \bm{z}_g') \vert \le \rho \Vert \bm{z}_g - \bm{z}_g' \Vert	
\label{eqlipschitz}.
\end{equation}
We denote $\bm{z}_i \sim N(\mu, \sigma)$ and $\bm{z}_g \sim p_g$, a certain continuous prior distribution. The Lipschitz of $G$ for $\bm{z}_g$ indicates that the condition we introduce can continuously control the generation. It is obvious that if $p_g$ is a discrete distribution, the model is altered into an ordinary cGAN and the Lipschitz of $G(\bm{z}_i, \bm{z}_g)$ for $\bm{z}_g$ turns false, because $\bm{z}_g$ does not exist in a metric space and $\Vert \bm{z}_g - \bm{z}_g' \Vert$ makes no sense anymore.

In GC-GAN, we aim to solve the problem in Eq.\eqref{problem}, where $\bm{z}_i$ and $\bm{z}_g$ represent identity information and facial expression condition of generated faces respectively. And $\bm{z}_i$ and $\bm{z}_g$ here are independent since the known image $\bm{I}^u_j$ and geometry vector $\bm{g}^v_k$ don't necessarily refer to the same subject. Compared to discrete distribution of facial expressions, if $p_g$ is some continuous distribution derived from facial landmarks, conditions $\bm{z}_g$ can guide the generation of facial expressions more precisely.

However, it increases the risk of errors to introduce geometry variables $\bm{g}^v_k$ into cGANs directly \textit{i.e.}, $\bm{z}_g = \bm{g}^v_k$. Original facial landmark space is badly separable for facial expressions. So when $\bm{g}^v_k$ changes as conditions, the changes of generated images will get out of control, \textit{i.e.}, the constant $\rho$ is too large or approximately infinity. It also has bad interpretability to concatenate landmark vector and prior in the input of generator. So an embedding method is necessary to transform the original geometry information into a more effective representation. Since the objective of contrastive learning is based on the calculation of metrics, we apply a contrastive learning network $E$ to the landmarks and define an embedding variable $\bm{z}_g = E_{enc}(\bm{g}^v_k)$. Through contrastive learning, landmarks are encoded into a semantic space of facial expression, which reduces the dimensions, extracts facial expression features, and provides the different features with better separability and appropriate distance. Thus, when the conditions $\bm{z}_g$ in semantic space are introduced and the prior $\bm{z}_i$ fixed, the generated images are Lipschitz continuous over the facial expression conditions with the identity information preserved.

\subsection{Learning strategy}

\paragraph{Geometry-contrastive learning}\label{section:contr}

Contrastive learning is conducted on facial geometry with reference facial landmarks, where the $xy$-coordinates of facial landmarks are arranged into a one-dimensional vector as input for the geometry embedding network. Landmark pairs $(\bm{g}^v_{k}, \bm{g}^{ref}_{(\cdot)})$ are prepared for training, in which $\bm{g}^v_{k}$ is the target expression while $\bm{g}^{ref}_{(\cdot)}$ is the reference expression. $(\cdot)$ represents a certain subject. After a transform function $E_{enc}(\cdot)$, the facial landmarks are mapped into the embedding space. Our goal is to measure the similarity between $E_{enc}(\bm{g}^v_{k})$ and $E_{enc}(\bm{g}^{ref}_{(\cdot)})$ according to their expression labels. The contrastive loss $L_{contr}$ is formulated as below:
\begin{equation}
\begin{split}
\min_{E_{enc}} L_{contr} = & \frac{\alpha}{2}\max \left(0, m- \left\|E_{enc}(\bm{g}^v_{k}) - E_{enc}(\bm{g}^{ref}_{(\cdot)}) \right\|^2_2 \right) \label{eqcontr} \\
& + \frac{1-\alpha}{2} \left\| E_{enc}(\bm{g}^v_{k})-E_{enc}(\bm{g}^{ref}_{(\cdot)}) \right\|^2_2,
\end{split}        
\end{equation}
where $\alpha = 0$, if the expression labels are the same, $\alpha = 1$ otherwise. $m$ is a margin which is always greater than 0. $L_{contr}$ enables our embedding network to focus on facial expression itself regardless of different subjects and plays an essential part for 
pushing facial landmarks to reside in a semantic manifold. The semantic manifold is explained in detail in Section \ref{section:effects_contr}.

\paragraph{Adversarial learning}\label{section:adv}

In our model, the adversarial losses for $G$ and $D$ are formulated as follows:
\begin{align}
	\label{eqadvd}
	\min_{D} L_{discr} &= -\operatorname*{\mathbb{E}}\limits_{\bm{I} \sim p_{\bm{I}}} D(\bm{I}^{u}_j) + \operatorname*{\mathbb{E}}\limits_{\bm{I} \sim p_{\bm{I}}, \bm{g} \sim p_{\bm{g}}} D(\hat{\bm{I}}^{v}_j), \\
	\label{eqadvg}
	\min_{G} L_{gen} &= -\operatorname*{\mathbb{E}}\limits_{\bm{I} \sim p_{\bm{I}}, \bm{g}\sim p_{\bm{g}}} D(\hat{\bm{I}}^{v}_j), \\
	\label{eqadv}
	\min_{G,D} L_{adv} &= L_{discr} + L_{gen},
\end{align}
where $p_{\bm{I}}$ and $p_{\bm{g}}$ indicate the distribution of real facial images and real facial landmarks, respectively. $\hat{\bm{I}}^{v}_j$ represents the generated image computed by $G_{dec}(G_{enc}(\bm{I}^u_j),  E_{enc}(\bm{g}^{v}_j))$. The adversarial losses are optimized via WGAN-GP \cite{Gulrajani17_WGANGP}.

\paragraph{Reconstruction learning} 
The $\ell_1$ loss and $\ell_2$ loss are introduced for the reconstruction of faces and landmarks, respectively. For the generated facial image $\hat{\bm{I}}^{v}_j$, we employ $\ell_1$ distance to compute a pixel-to-pixel difference between generated image and target real image, which is formulated as $
\min_{G} L_{ir} = \left\|\bm{I}^v_j-\hat{\bm{I}}^{v}_j \right\|_1
\label{eqrecI}$.
The mixture of adversarial loss and $L_{ir}$ accelerate the convergence and enhance image visual quality. For facial landmarks, $\ell_2$ distance is employed to compute the difference between
reconstructed landmarks and input landmarks, which is formulated as
$\min_{E} L_{gr} = \left\|\bm{g}^{v}_k-\hat{\bm{g}}^{v}_k \right\|^2_2.
\label{eqrecL}$ Reconstruction learning enables latent vector $\bm{z}$ to preserve enough information for the reconstruction of inputs.

\paragraph{Overall learning} 
First, network $E$ is pretrained on the training set. The loss function is $\min_{E} L_{E} =\lambda_1 L_{contr} + \lambda_2 L_{gr}
\label{eqfull1}$. Then network $G$ and $D$ are trained together while the parameters of network $E$ are fixed. The loss function is $\min_{G,D} L_{G} = \lambda_3 L_{ir} + \lambda_4 L_{adv}\label{eqfull2}$. $\lambda_1$, $\lambda_2$, $\lambda_3$ and $\lambda_4$ are set to $1$, $1$, $1$ and $10^{-3}$, respectively. The detailed parameters of GC-GAN are provided in Appendix-A.

\section{Experiments}

\subsection{Datasets and protocols}

\noindent To evaluate the proposed GC-GAN, experiments have been conducted on four popular facial datasets: Multi-PIE \cite{Gross10_mpie}, CK+ \cite{Lucey10_ckplus}, BU-4DFE \cite{Yin08_BU4D} and CelebA \cite{Liu15_celeba}.

Multi-PIE consists of 754,200 images from 337 subjects with large variations in head pose, illumination, and facial expression. We select a subset of images with three head poses ($0^\circ, \pm 15^\circ$), 20 illumination conditions, and all six expressions.
CK+ is composed of 327 image sequences from 118 subjects with seven prototypical emotion labels. Each sequence starts with a neutral emotion and ends with the peak of a certain emotion.
BU-4DFE contains 101 subjects, each one displaying six acted categorical facial expressions with moderate head pose variations. Since this dataset does not contain frame-wise facial expression, we manually select neutral and apex frames of each sequence. 
CelebA is a large-scale facial dataset, containing  202,599 face images of celebrities. All experiments are performed in a person-independent way, the training set and test set for each dataset are split based on subjects with proportions of 90\% and 10\%, respectively.

To prepare the training triplet $(\bm{I}^{u}_j, \bm{g}^{v}_k, \bm{g}^{ref}_{(\cdot)})$, we create all triplets of facial expressions per identity. $\bm{g}^{v}_k$ is taken from $\bm{I}^{v}_k$ and $\bm{g}^{ref}_{(\cdot)}$ is generated by sampling at random. Note that $\bm{g}^{ref}_{(\cdot)}$ and facial expression labels are not required in test. In our experiments, 68 facial landmark points are obtained through dlib \footnote{http://dlib.net/} which implements the method proposed in \cite{Kazemi14_ERT}. Since faces in our experiments remain near-frontal, the accuracy of facial landmark detection is over 95\% when dlib is used, which is sufficient for our work.
The facial landmarks include points of two eyebrows, two eyes, the nose, the lips, and the jaw. The facial images are aligned according to inner eyes and bottom lip. Then, face regions are cropped and resized into $64 \times 64$. Pixel value of images and $xy$-coordinates of facial landmarks are both normalized into $[-1,1]$. 

\subsection{Facial expression generation} 

\begin{figure}
	\centering
	\begin{minipage}{12cm}
		\centering
		\subfigure{\includegraphics[width=12cm]{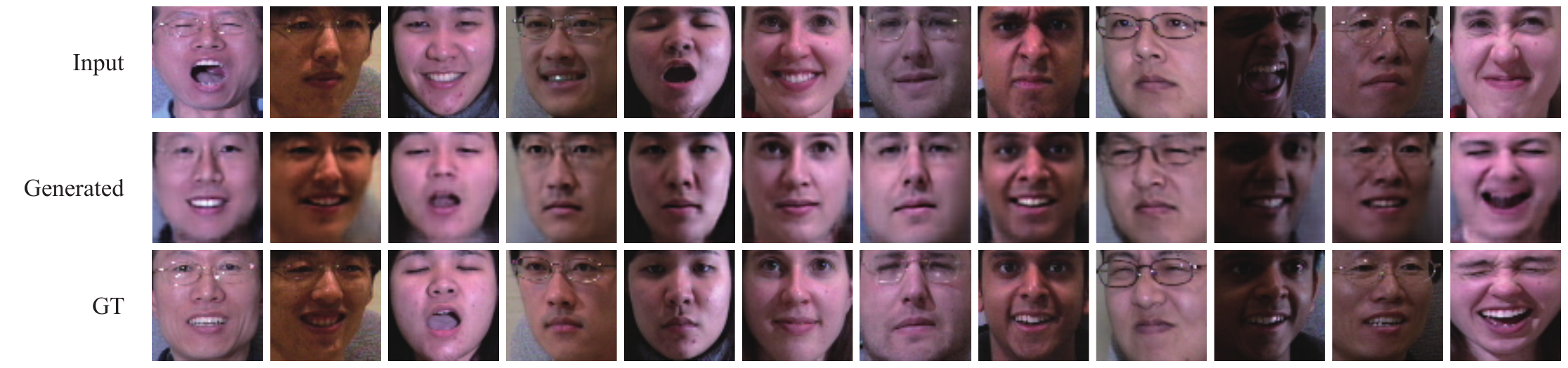}}
		\footnotesize{(a) Multi-PIE}
	\end{minipage}\\[1mm]
	\begin{minipage}{12cm}
		\centering
		\subfigure{\includegraphics[width=12cm]{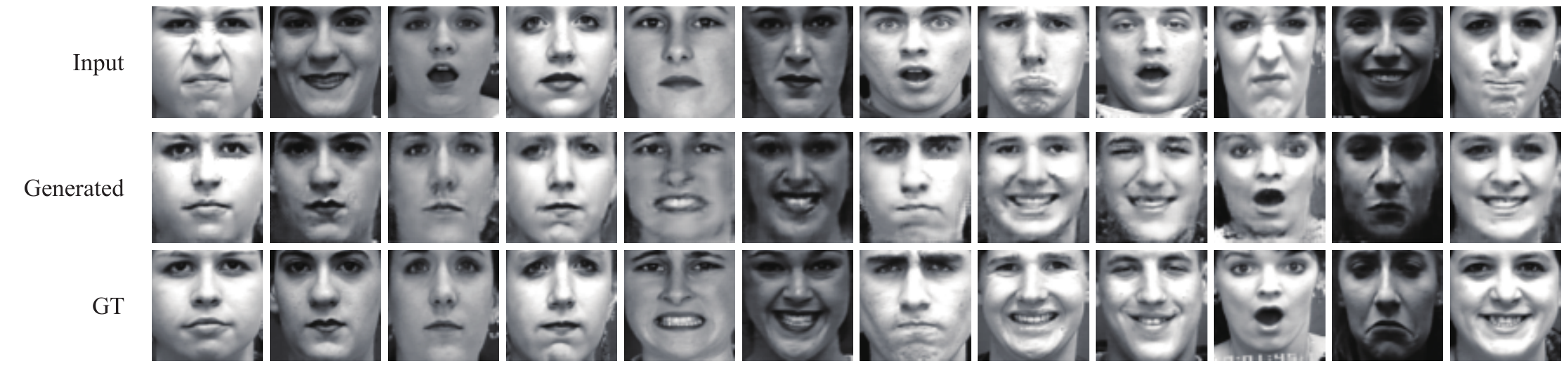}}
		\footnotesize{(b) CK+}
	\end{minipage}\\[1mm]
	\begin{minipage}{12cm}
		\centering
		\subfigure{\includegraphics[width=12cm]{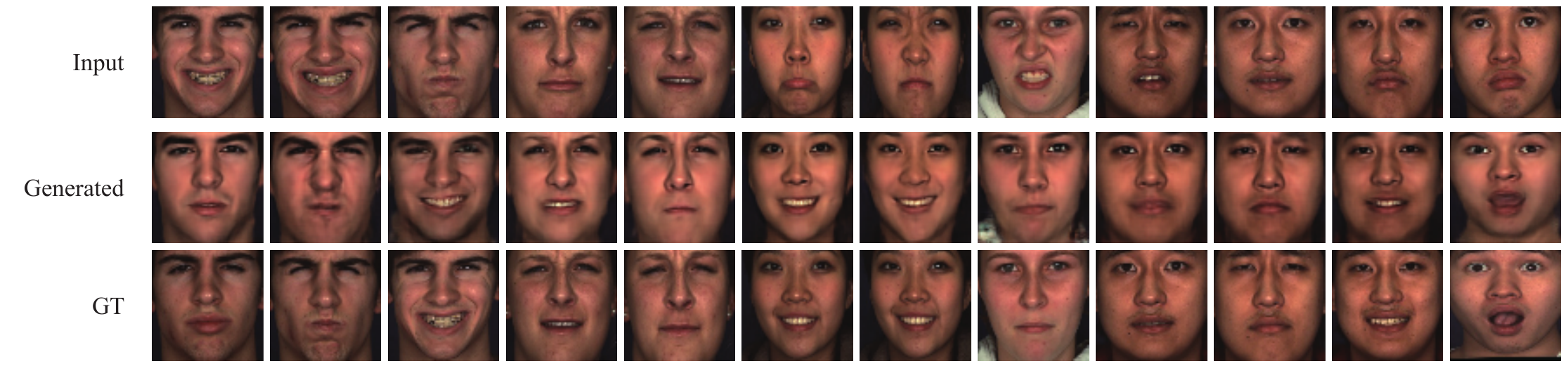}}
		\footnotesize{(c) BU-4DFE}
	\end{minipage}
	\caption{Results of facial expression synthesis on (a) Multi-PIE, (b) CK+, and (c) BU-4DFE. Three images from top to bottom in each column represent the input image, generated image, and ground truth, respectively.}
	\label{fig_synth}
\end{figure}

In order to evaluate whether the synthetic faces are generated by the guidance of the target facial expression, we qualitatively and quantitatively compare the generated faces with the ground truth. The results are shown in Figure \ref{fig_synth}, in which the generated faces are identity-preserving and are similar to the target facial expressions. For quantitative comparison, SSIM (structural similarity index measure)  and PSNR (peak-signal-to-noise ratio) are used as two evaluation metrics. The detailed comparison results are shown in Table \ref{table_comparison}. We conduct ablation study to evaluate contributions of three losses $L_{ir}$, $L_{gr}$, and $L_{contr}$, respectively. According to Table \ref{table_comparison}, GC-GAN performs better when supervised by the three losses. We compare our method with CDAAE \cite{Zhou17_FES}, which uses one-hot codes to represent facial expressions. As shown in Table \ref{table_comparison}, our method substantially outperforms CDAAE on both measures across all the datasets, indicating that the images generated using our method are more similar to the ground truth aided by the embedded geometry features. The experiments mentioned above demonstrate that our proposed GC-GAN effectively incorporates geometry information in conditioned facial expression generation.

\begin{table}[t]
	\caption{Comparison with other settings and method on Multi-PIE, CK+ and BU-4DFE in terms of SSIM and PSNR.}
	\label{table_comparison}
	\centering
	\begin{tabular}{lcccccc}
		\toprule
		\multicolumn{1}{c}{\multirow{2}{*}{}} & \multicolumn{3}{c}{SSIM}    & \multicolumn{3}{c}{PSNR}   \\ 
		\cmidrule{2-7} 
		\multicolumn{1}{c}{}                  & \multicolumn{1}{c}{Multi-PIE} & \multicolumn{1}{c}{CK+} & \multicolumn{1}{c}{BU-4DFE} & \multicolumn{1}{c}{Multi-PIE}
		& \multicolumn{1}{c}{CK+} & \multicolumn{1}{c}{BU-4DFE} \\ \midrule
		\multicolumn{1}{c}{GC-GAN (ours)}                   & \multicolumn{1}{c}{\textbf{0.687}}                     & \multicolumn{1}{c}{\textbf{0.769}}       & \multicolumn{1}{c}{\textbf{0.725}}             &   \multicolumn{1}{c}{\textbf{26.731}}                   & \multicolumn{1}{c}{\textbf{27.665}}       & \multicolumn{1}{c}{\textbf{26.915}} \\ 
		\quad w/o $L_{ir}$      & 0.663     & 0.756     & 0.702     & 25.917     & 26.781   & 25.136   \\ 
		\quad w/o $L_{gr}$      & 0.682     & 0.768     & 0.722     & 26.701     & 27.463   & 26.875   \\ 
		\quad w/o $L_{contr}$   & 0.675     & 0.763     & 0.705         & 26.433     & 27.325   & 26.325    \\ 
		CDAAE \cite{Zhou17_FES}         & 0.669     & 0.765     & 0.710         & 26.371     & 26.973   & 26.122    \\
		\bottomrule
	\end{tabular}
\end{table}



\subsection{Analysis of geometry-contrastive learning}\label{section:effects_contr}

\begin{figure}
	\centering
	\begin{minipage}{6.5cm}
		\centering
		\subfigure{\includegraphics[width=6.5cm]{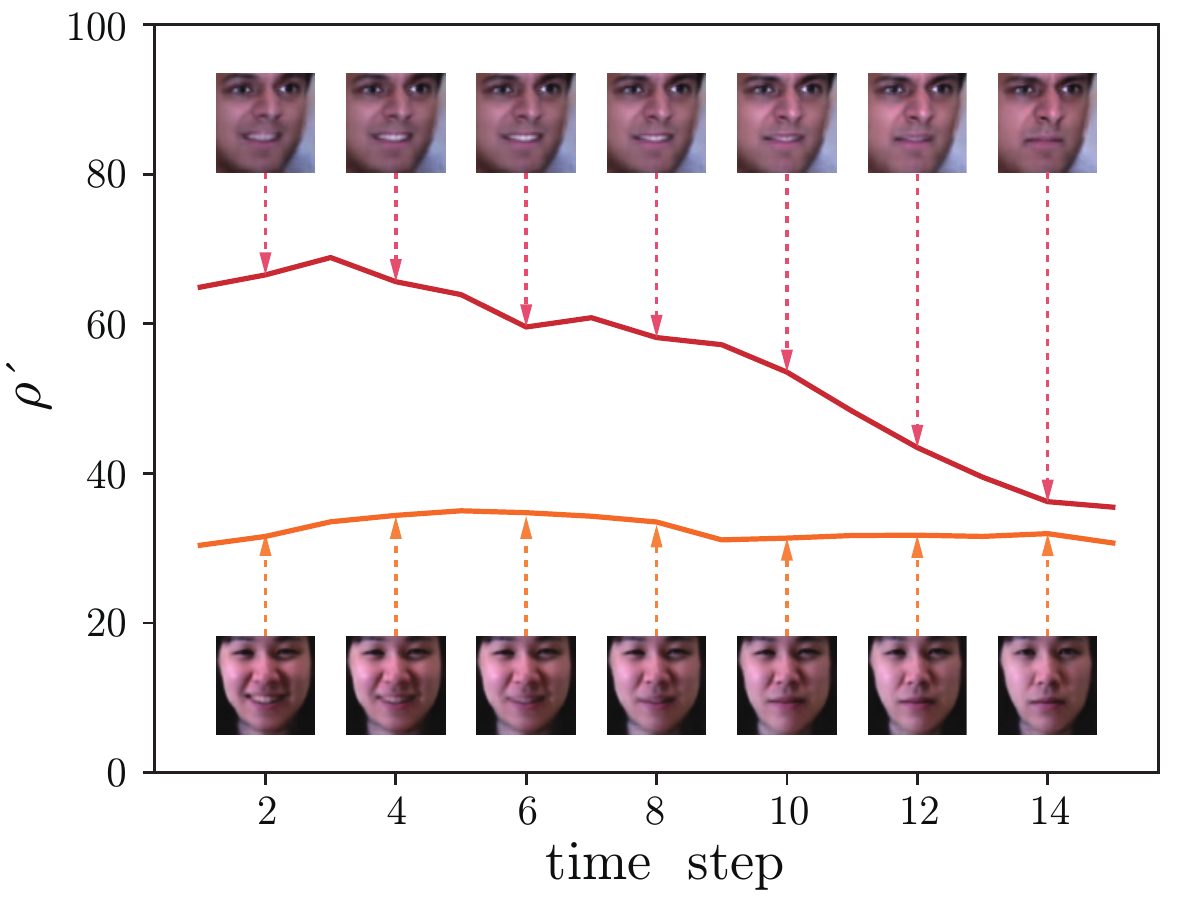}}\\
		\footnotesize{(a)}
	\end{minipage}\quad
	\begin{minipage}{6.5cm}
		\centering
		\subfigure{\includegraphics[width=6.5cm]{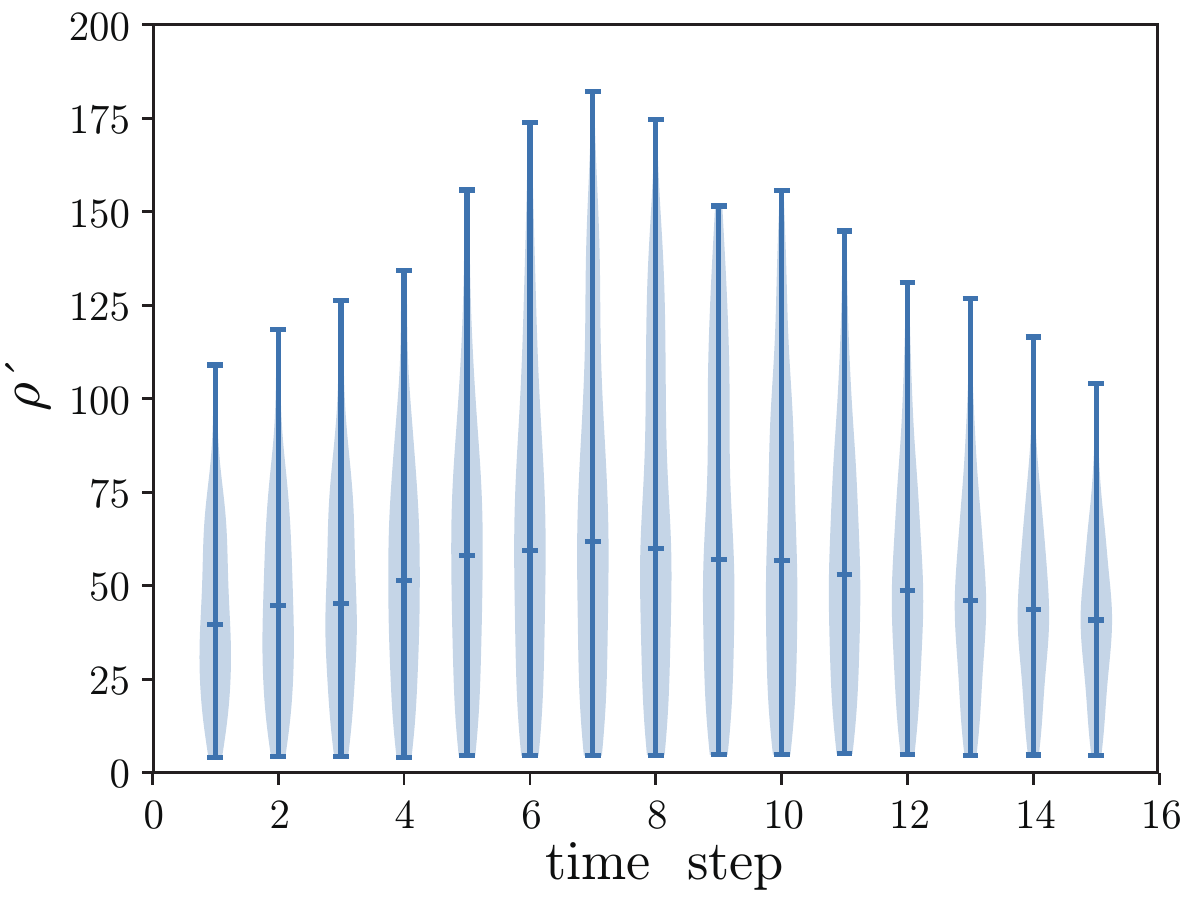}}\\
		\footnotesize{(b)}
	\end{minipage}
	\caption{Evaluation of Lipschitz continuity. (a) shows two generated facial expression sequences from \textit{smile} to \textit{disgust} in red curve and from \textit{smile} to \textit{neutral} in orange curve, respectively. (b) shows the violin plot at each time step computed over the whole test set.}
	\label{fig_lipchitz}
	\vskip-1mm
\end{figure}

First, we evaluate the Lipschitz continuity of generated faces $G(\bm{z}_i, \bm{z}_g)$ over facial expression conditions $\bm{z}_g$. For interpolation of $\bm{z}_g$, we would like to seek a sequence of $N$ images with a smooth transition between two different facial expressions of the same person (e.g. $\bm{I}^{u}_j$ and $\bm{I}^{v}_j$). In our case, we conduct equal interval sampling $\left[\frac{t}{N} \cdot E_{enc}(\bm{g}^u_j)+(1-\frac{t}{N}) \cdot E_{enc}(\bm{g}^v_j)\right]_{t=0}^N$ on $\bm{z}_g$. Some generated facial expression sequences are shown in Figure \ref{fig_lipchitz}(a). According to Eq.\eqref{eqlipschitz}, we compute $\rho'=\frac{\vert G(z_i, z_g) - G(z_i, z_g') \vert}{\Vert z_g - z_g' \Vert}$ between two adjacent frames over the whole test set composed of 58,724 pairs from 33 subjects, and the violin plot is shown in Figure \ref{fig_lipchitz}(b), where the mean at each time step is around 50 and the upper bound also exists. Therefore, the Lipschitz constant $\rho$ in Eq.\eqref{eqlipschitz} is approximately found.
 
\begin{figure}[t]
	\centering
	\begin{minipage}{4cm}
		\centering
		\subfigure{\includegraphics[width=4cm]{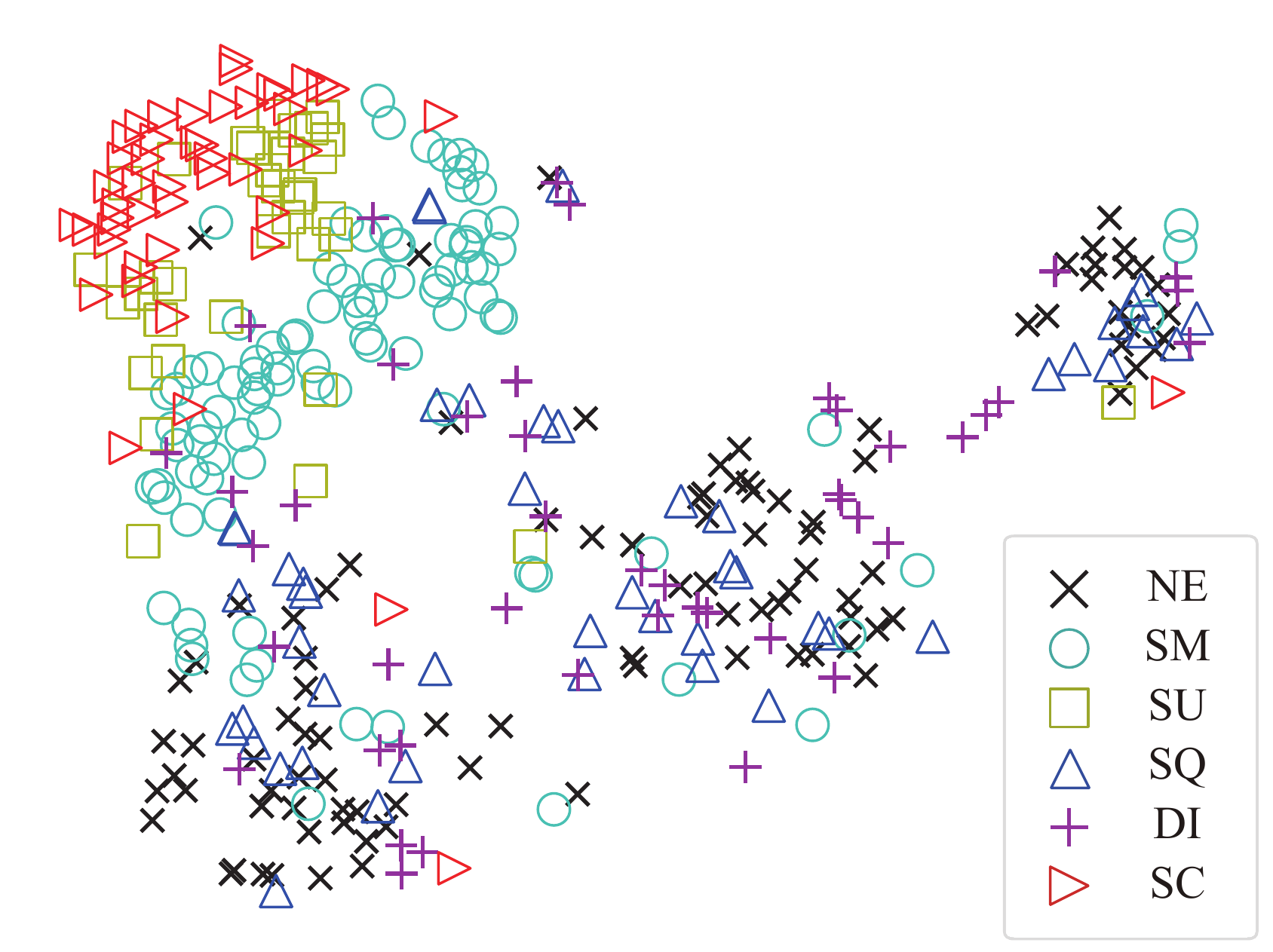}}\\
		\footnotesize{(a)}
	\end{minipage}\quad
	\begin{minipage}{4cm}
		\centering
		\subfigure{\includegraphics[width=4cm]{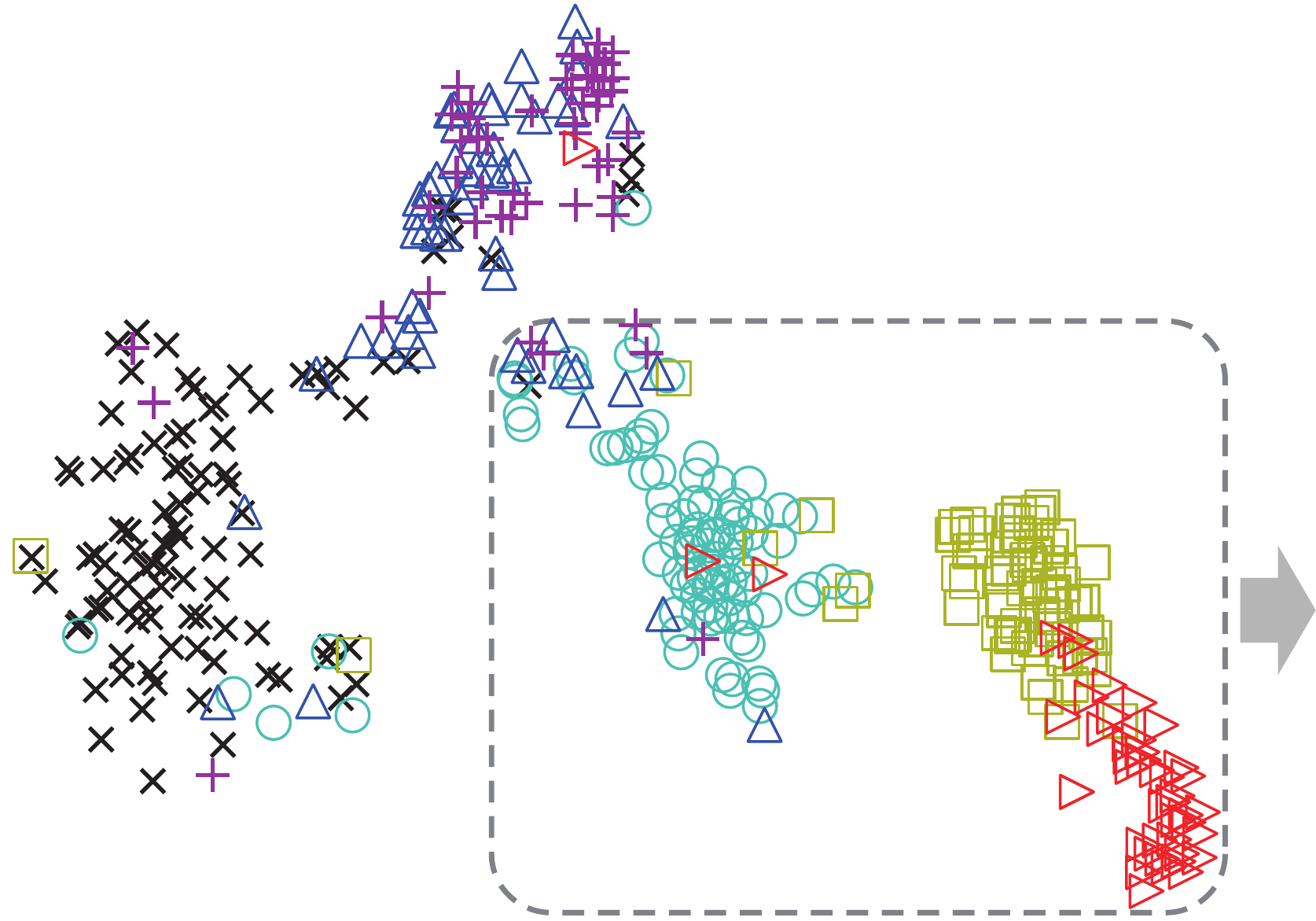}}\\
		\footnotesize{(b)}
	\end{minipage}
	\begin{minipage}{5.5cm}
		\centering
		\subfigure{\includegraphics[width=5.5cm]{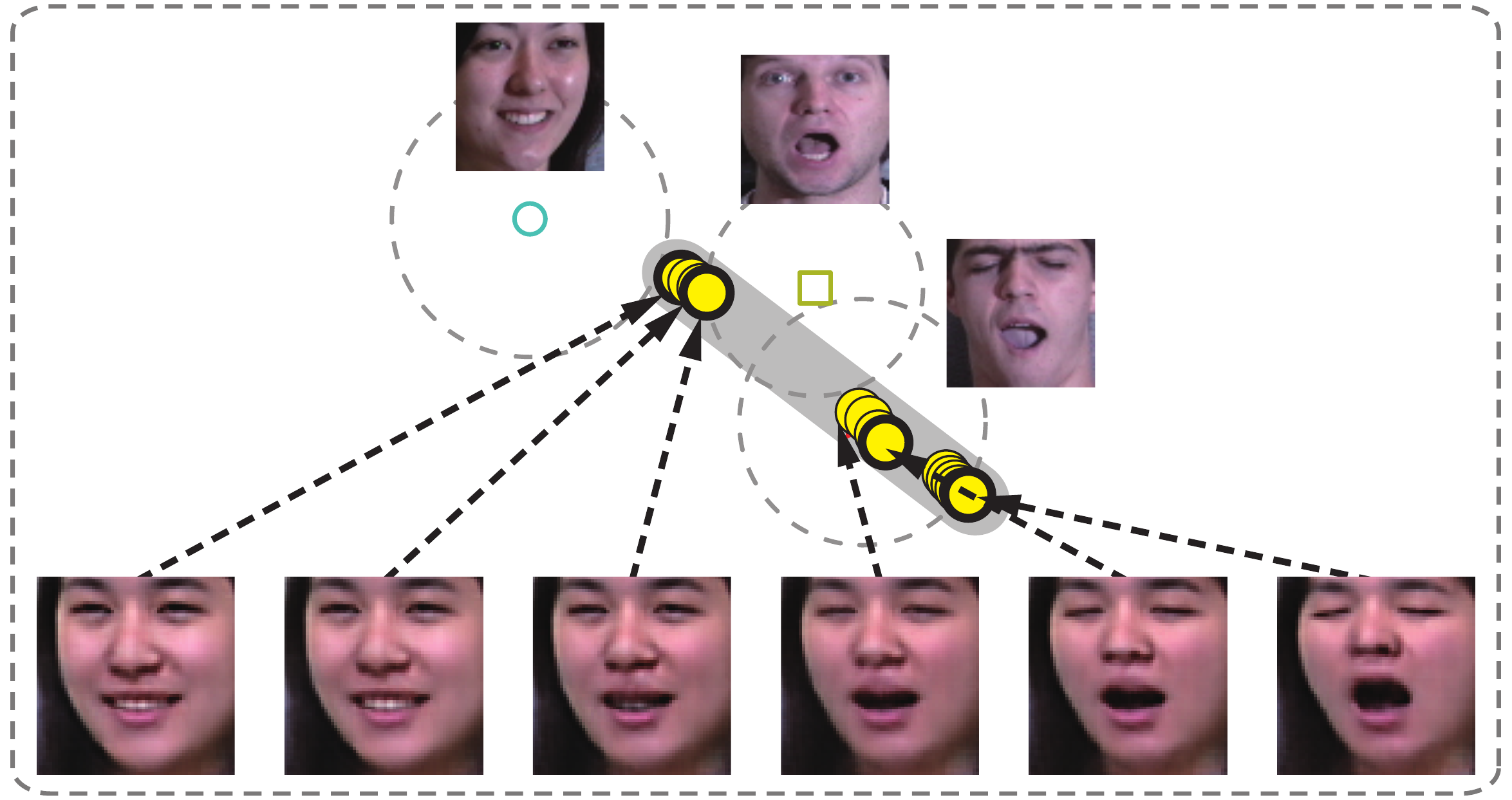}}\\
		\footnotesize{(c)}
	\end{minipage}	
	\caption{Visualization of the embedded manifolds. The manifold of (a) GC-GAN (w/o $L_{contr}$), and the manifold of (b) GC-GAN. (c) The embedding points of a generated facial expression sequence.}
	\label{fig_manifold_exp_tsne}
	\vskip-1mm
\end{figure}

In order to evaluate whether the manifold of $\bm{z}_g$ is semantic-aware, we visualize the embedding manifold of facial landmarks by t-SNE \cite{Maaten08_tSNE}. For the purpose of evaluating the effectiveness of contrastive learning, we train our model without contrastive loss $L_{contr}$. The corresponding manifold of facial landmarks is shown in Figure \ref{fig_manifold_exp_tsne}(a), in which the embedding features cannot be clustered properly without the guidance of contrastive learning. The manifold of GC-GAN is shown in Figure \ref{fig_manifold_exp_tsne}(b) where the embedding features are clustered according to their emotion states. It can be seen that the data scatters of \textit{disgust} and \textit{squint} are mixed up with each other due to the intrinsic geometry similarity between the two facial expressions. 
In addition, we present an embedding line consisting of a sequence of points corresponding to the transition of facial expressions from \textit{smile} to \textit{scream} of the same person obtained by equal interval sampling on facial landmarks, which is shown in Figure \ref{fig_manifold_exp_tsne}(c). It can be observed that the eyes are closing and the mouth is opening from \textit{smile} to \textit{scream}. So there could exist some points with open eyes and open mouth simultaneously which is exactly the characteristic of \textit{surprise}, and could account for the reason why the embedding line comes across the region of \textit{surprise}. The results indicate that the manifold of $\bm{z}_g$ is both continuous and semantic-aware for GC-GAN.

%

\begin{figure}
	\centering
	\includegraphics[width=14cm]{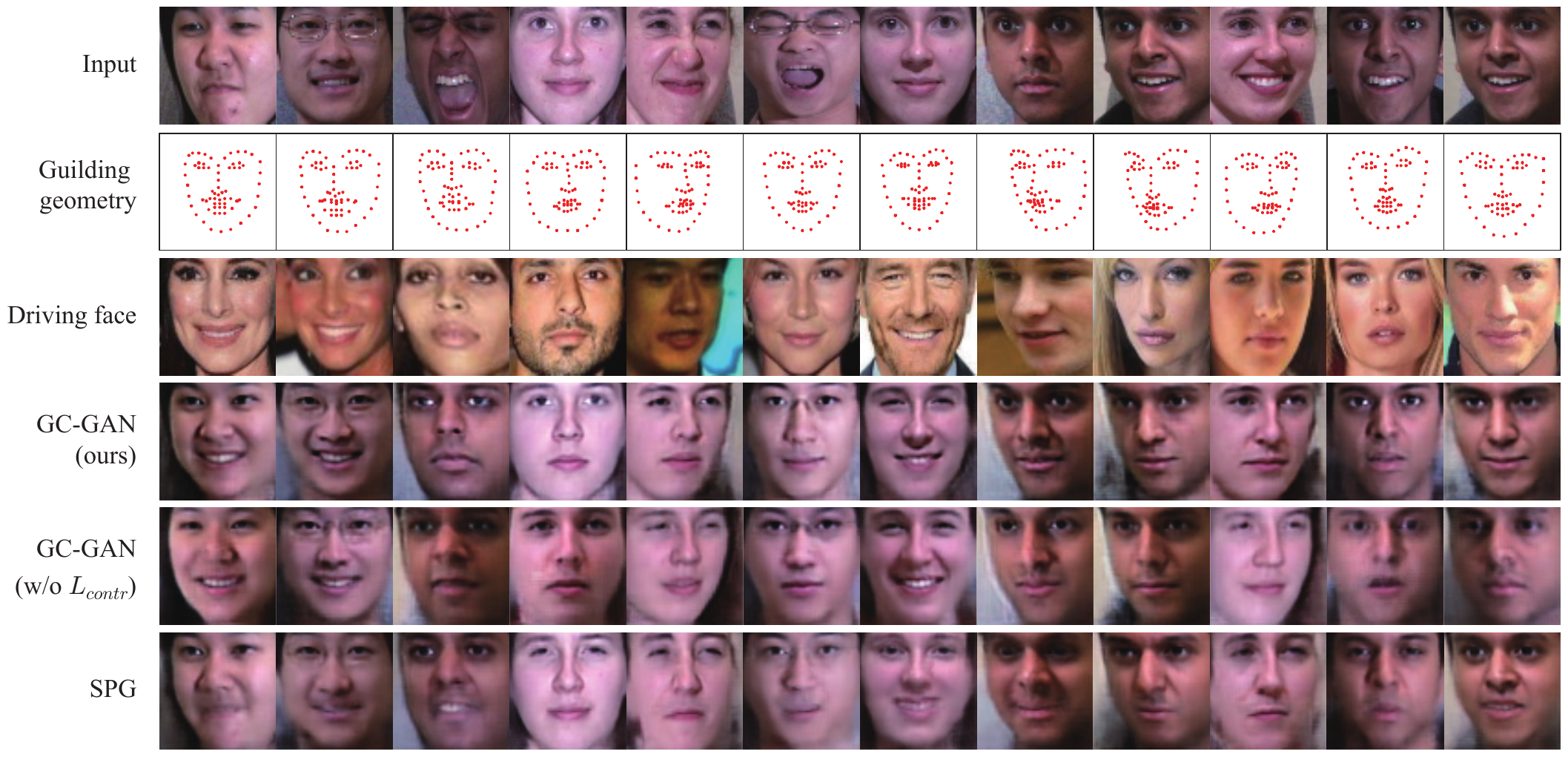}
	\caption{Results of facial expression transfer on Multi-PIE from CelebA. Input faces are selected from Multi-PIE, and guiding facial geometries are taken from the corresponding driving faces in CelebA.}
	\label{fig_synth_celeba}
\end{figure}

\begin{table}[t]
	\caption{Evaluation results of facial expression transfer on Multi-PIE from CelebA.}
	\label{table_eval_on_celeba}
	\centering
	\begin{tabular}{lcc}
		\toprule
		& Accuracy of face verification & Similarity of facial expression \\
		\midrule
		GC-GAN (ours)	& \textbf{0.977} & \textbf{0.896} \\
		GC-GAN (w/o $L_{contr}$)	& 0.942 & 0.880 \\
		SPG \cite{Ma17_PG,Song17_G2GAN}	& 0.875 & 0.854 \\
		\bottomrule
	\end{tabular}
	\vskip-4mm
\end{table}

\subsection{Facial expression transfer}\label{section:transfer}

\noindent To further evaluate the generality of our proposed method, transferring a person's emotion to different emotions of another person directly is also tested. Facial expression transfer between different persons are difficult due to the individual variances, especially in the absence of neutral faces for reference.

In our case, the target expressions are taken from other people of different datasets. Here, we randomly select 1000 facial images from CelebA. The facial landmarks of these images serve as guiding landmarks. Note that there are only six basic emotion states in Multi-PIE while the emotion states in CelebA are various and spontaneous. Figure \ref{fig_synth_celeba} shows some representative samples from our experiment, in which the generated faces are identity-preserving and exhibit similar expressions with driving faces even there exist big differences in face shape between input faces and driving faces. 

For comparison, a more general configuration SPG is adopted, which is similar to PG$^2$ \cite{Ma17_PG} and G2GAN \cite{Song17_G2GAN}. In detail, we treat the guiding landmarks as an additional feature map and directly concatenate it with the input face, and train our model in this configuration. The corresponding results of SPG are shown in the lowest line of Figure \ref{fig_synth_celeba}. 

Since there don't exist ground truth faces corresponding to generated faces, SSIM and PSNR are not suitable for quantitative evaluation.
In order to validate whether the generated faces are identity-preserving, a pretrained face identification model VGG-Face \cite{Parkhi15_vggface} is adopted to conduct face verification on the generated faces.
In detail, we extract the output of the last convolution layer for measuring the identity similarity between two faces. Two faces are considered as matched if their cosine similarity is no less than 0.65. In order to evaluate the similarity of facial expressions between generated and driving faces, the cosine similarity of $\bm{z}_g$ of facial landmarks between generated and driving faces is adopted.
The accuracy of face verification and the mean similarity of facial expressions are shown in Table \ref{table_eval_on_celeba}.
Compared with our method, SPG is not able to capture the target facial expression of driving faces and brings extra artifact, indicating that the model under this configuration cannot handle the misalignments between input faces and driving facial landmarks properly.
Our model shows the promising capability of manipulating facial expressions while preserving identity information.


\section {Conclusion}

\noindent In this paper, we have proposed a Geometry-Contrastive Generative Adversarial Network (GC-GAN) for transferring facial expressions across different persons. Through contrastive learning, facial expressions are pushed to reside onto a continuous semantic-aware manifold. Benefited from this manifold, pixel-wise misalignments between different subjects are weakened and identity-preserving faces with more expression characteristics are generated. Experimental results demonstrate that there exists semantic consistency between the latent space of facial expressions and the image space of faces.




\small

\bibliographystyle{plain}
\bibliography{nips18}

\begin{thebibliography}{10}

\bibitem{Averbuch17_portraits}
Hadar Averbuch-Elor, Daniel Cohen-Or, Johannes Kopf, and Michael~F Cohen.
\newblock Bringing portraits to life.
\newblock {\em {ACM} Trans. Graph}, 36(6):196:1--196:13, 2017.

\bibitem{Balakrishnan18_synthesizePose}
Guha Balakrishnan, Amy Zhao, Adrian~V Dalca, Fredo Durand, and John Guttag.
\newblock Synthesizing images of humans in unseen poses.
\newblock In {\em CVPR (To appear)}, 2018.

\bibitem{Choi17_StarGAN}
Yunjey Choi, Minje Choi, Munyoung Kim, Jung-Woo Ha, Sunghun Kim, and Jaegul
  Choo.
\newblock Stargan: Unified generative adversarial networks for multi-domain
  image-to-image translation.
\newblock In {\em CVPR (To appear)}, 2018.

\bibitem{Chopra05_contrastive}
Sumit Chopra, Raia Hadsell, and Yann LeCun.
\newblock Learning a similarity metric discriminatively, with application to
  face verification.
\newblock In {\em CVPR}, pages 539--546, 2005.

\bibitem{Dale11_reenactment}
Kevin Dale, Kalyan Sunkavalli, Micah~K Johnson, Daniel Vlasic, Wojciech
  Matusik, and Hanspeter Pfister.
\newblock Video face replacement.
\newblock {\em {ACM} Trans. Graph}, 30(6):130:1--130:10, 2011.

\bibitem{Hui18_ExprGAN}
Hui Ding, Kumar Sricharan, and Rama Chellappa.
\newblock Exprgan: Facial expression editing with controllable expression
  intensity.
\newblock In {\em AAAI (To appear)}, 2018.

\bibitem{Ekman77_FACS}
Paul Ekman and Wallace~V Friesen.
\newblock Facial action coding system.
\newblock {\em Consulting Psychologists Press, Stanford University}, 1977.

\bibitem{Goodfellow14_GAN}
Ian Goodfellow, Jean Pouget-Abadie, Mehdi Mirza, Bing Xu, David Warde-Farley,
  Sherjil Ozair, Aaron Courville, and Yoshua Bengio.
\newblock Generative adversarial nets.
\newblock In {\em NIPS}, pages 2672--2680, 2014.

\bibitem{Gross10_mpie}
Ralph Gross, Iain Matthews, Jeffrey Cohn, Takeo Kanade, and Simon Baker.
\newblock Multi-{PIE}.
\newblock {\em Image and Vision Computing}, 28(5):807--813, 2010.

\bibitem{Gulrajani17_WGANGP}
Ishaan Gulrajani, Faruk Ahmed, Martin Arjovsky, Vincent Dumoulin, and Aaron
  Courville.
\newblock Improved training of wasserstein gans.
\newblock In {\em NIPS}, pages 5769--5779, 2017.

\bibitem{Qi_LSGAN}
Qi~Guojun.
\newblock Loss-sensitive generative adversarial networks on lipschitz
  densities.
\newblock {\em arXiv preprint arXiv: 1701.06264}, 2017.

\bibitem{Kazemi14_ERT}
Vahid Kazemi and Sullivan Josephine.
\newblock One millisecond face alignment with an ensemble of regression trees.
\newblock In {\em CVPR}, pages 1867--1874, 2014.

\bibitem{Kossaifi17_GAGAN}
Jean Kossaifi, Linh Tran, Yannis Panagakis, and Maja Pantic.
\newblock {GAGAN}: {G}eometry-aware generative adversarial networks.
\newblock In {\em CVPR (To appear)}, 2018.

\bibitem{Larsen16_VAEGAN}
Anders Boesen~Lindbo Larsen, S{\o}ren~Kaae S{\o}nderby, Hugo Larochelle, and
  Ole Winther.
\newblock Autoencoding beyond pixels using a learned similarity metric.
\newblock In {\em ICML}, pages 1558--1566, 2016.

\bibitem{Liu15_celeba}
Ziwei Liu, Ping Luo, Xiaogang Wang, and Xiaoou Tang.
\newblock Deep learning face attributes in the wild.
\newblock In {\em ICCV}, pages 3730--3738, 2015.

\bibitem{Lucey10_ckplus}
Patrick Lucey, Jeffrey~F Cohn, Takeo Kanade, Jason Saragih, Zara Ambadar, and
  Iain Matthews.
\newblock The extended {C}ohn-{K}anade dataset ({CK}+): {A} complete dataset
  for action unit and emotion-specified expression.
\newblock In {\em CVPRW}, pages 94--101, 2010.

\bibitem{Ma17_PG}
Liqian Ma, Xu~Jia, Qianru Sun, Bernt Schiele, Tinne Tuytelaars, and Luc
  Van~Gool.
\newblock Pose guided person image generation.
\newblock In {\em NIPS}, pages 405--415, 2017.

\bibitem{Maaten08_tSNE}
Laurens van~der Maaten and Geoffrey Hinton.
\newblock Visualizing high-dimensional data using t-sne.
\newblock {\em JMLR}, 9(2):2579--2605, 2008.

\bibitem{Mirza14_cGAN}
Mehdi Mirza and Simon Osindero.
\newblock Conditional generative adversarial nets.
\newblock {\em arXiv preprint arXiv:1411.1784}, 2014.

\bibitem{Parkhi15_vggface}
Omkar~M Parkhi, Andrea Vedaldi, Andrew Zisserman, et~al.
\newblock Deep face recognition.
\newblock In {\em BMVC}, pages 41.1--41.12, 2015.

\bibitem{Reed16_draw}
Scott~E Reed, Zeynep Akata, Santosh Mohan, Samuel Tenka, Bernt Schiele, and
  Honglak Lee.
\newblock Learning what and where to draw.
\newblock In {\em NIPS}, pages 217--225, 2016.

\bibitem{Song17_G2GAN}
Lingxiao Song, Zhihe Lu, Ran He, Zhenan Sun, and Tieniu Tan.
\newblock Geometry guided adversarial facial expression synthesis.
\newblock {\em arXiv preprint arXiv: 1712.03474}, 2017.

\bibitem{Thies15_realtime}
Justus Thies, Michael Zollh{\"o}fer, Matthias Nie{\ss}ner, Levi Valgaerts, Marc
  Stamminger, and Christian Theobalt.
\newblock Real-time expression transfer for facial reenactment.
\newblock {\em {ACM} Trans. Graph}, 34(6):183--1, 2015.

\bibitem{Thies16_reenactment}
Justus Thies, Michael Zollhofer, Marc Stamminger, Christian Theobalt, and
  Matthias Nie{\ss}ner.
\newblock Face2face: Real-time face capture and reenactment of rgb videos.
\newblock In {\em CVPR}, pages 2387--2395, 2016.

\bibitem{Veit17_CSN}
Andreas Veit, Serge Belongie, and Theofanis Karaletsos.
\newblock Conditional similarity networks.
\newblock In {\em CVPR}, pages 1781--1789, 2017.

\bibitem{Wang17_metric}
Jian Wang, Feng Zhou, Shilei Wen, Xiao Liu, and Yuanqing Lin.
\newblock Deep metric learning with angular loss.
\newblock In {\em CVPR}, pages 2593--2601, 2017.

\bibitem{Wang18_every}
Wei Wang, Xavier Alameda-Pineda, Dan Xu, Pascal Fua, Elisa Ricci, and Nicu
  Sebe.
\newblock Every smile is unique: Landmark-guided diverse smile generation.
\newblock In {\em CVPR (To appear)}, 2018.

\bibitem{Yin08_BU4D}
Lijun Yin, Xiaochen Chen, Yi~Sun, Tony Worm, and Michael Reale.
\newblock A high-resolution 3d dynamic facial expression database.
\newblock In {\em FG}, pages 1--6, 2008.

\bibitem{Yin17_FFGAN}
Xi~Yin, Xiang Yu, Kihyuk Sohn, Xiaoming Liu, and Manmohan Chandraker.
\newblock Towards large-pose face frontalization in the wild.
\newblock In {\em ICCV}, pages 4010--4019, 2017.

\bibitem{Zeiler11_rbm}
Matthew~D Zeiler, Graham~W Taylor, Leonid Sigal, Iain Matthews, and Rob Fergus.
\newblock Facial expression transfer with input-output temporal restricted
  boltzmann machines.
\newblock In {\em NIPS}, pages 1629--1637, 2011.

\bibitem{Zhang17_fer}
Kaihao Zhang, Yongzhen Huang, Yong Du, and Liang Wang.
\newblock Facial expression recognition based on deep evolutional
  spatial-temporal networks.
\newblock {\em {IEEE} Trans. Image Processing}, 26(9):4193--4203, 2017.

\bibitem{Zhang17_age}
Zhifei Zhang, Yang Song, and Hairong Qi.
\newblock Age progression/regression by conditional adversarial autoencoder.
\newblock In {\em CVPR}, pages 4352--4360, 2017.

\bibitem{Zhou17_FES}
Yuqian Zhou and Bertram~Emil Shi.
\newblock Photorealistic facial expression synthesis by the conditional
  difference adversarial autoencoder.
\newblock In {\em ACII}, pages 370--376, 2017.

\end{thebibliography}

%
%
%

\section*{Appendix}

\section{Network architecture}

The detailed architectures of GC-GAN are shown in Table \ref{table_network_details}. In detail, generator network $G$ has a similar architecture to U-Net, but we only add one skip connection between the output of the first convolution layer and the output of the penultimate deconvolution layer, which enables $G$ to reuse low-level facial features related with identity information. For discriminator network $D$, the output is a $2\times2$ feature map. \\

\noindent  In training time, the margin $m$ in $L_{contr}$ is set to 5. All the networks $E$, $G$, and $D$ are optimized by Adam optimizer with $\beta_1=0.5$ and $\beta_2=0.999$. The batch size is 64 and the initial learning rate is set to $3 \times 10^{-4}$. We implement GC-GAN using Tensorflow.

\begin{table}[!h]
	\renewcommand\arraystretch{1.2}
	\centering
	\caption{The detailed architecture of \textit{GC-GAN}. In (\textit{De})\textit{Conv}($d$,$k$,$s$), $d$, $k$, and $s$ stand for the number of filters, the kernel size, and the stride, respectively. \textit{FC} represents a fully-connected layer. \textit{BN} is batch normalization and \textit{LN} is layer normalization. \textit{LReLU} refers to \textit{leaky ReLU}.}
	\vskip4mm
	\label{table_network_details}
	\begin{tabular}{lll}
		\hline
		\textbf{Embedding} $E$    & \textbf{Generator} $G$          & \textbf{Discriminator} $D$    \\ \hline
		FC(128), BN, ReLU                & Conv(64,5,2), BN, ReLU           & Conv(64,5,2), LN, LReLU 		  \\  
		FC(64), BN, ReLU                  & Conv(128,5,2), BN, ReLU         & Conv(128,5,2), LN, LReLU       \\ 
		FC(32), BN, ReLU                  & Conv(256,5,2), BN, ReLU         & Conv(256,5,2), LN, LReLU       \\ 
		FC(64), BN, ReLU                  & FC(1024), BN, ReLU                  & Conv(512,5,2), LN, LReLU       \\ 
		FC(128), BN, ReLU                & FC(128), BN, ReLU                    & Conv(1,5,2)							      \\ 
		FC(136), Tanh                       & FC(4096), BN, ReLU                   &                                                        \\
		                                                  & DeConv(256,5,2), BN, ReLU    &                                                        \\
		                                                  & DeConv(128,5,2), BN, ReLU    &                                                        \\ 
		                                                  & DeConv(64,5,2), BN, ReLU      &                                                        \\ 
		                                                  & DeConv(64,5,2), BN, ReLU      &                                                        \\ 
		                                                  & Conv(3,5,1), BN, Tanh              &                                                        \\ \hline
	\end{tabular}
\end{table}

\clearpage 
\section{Reviews}
\textbf{Meta-Review}

The paper of expression transfer in face analyses is very relevant. Although reviewers agree that including geometry in transfer with GANs and the analyses with contrastive loss and Lipschitz continuous semantic latent space are interesting, there is a generalized opinion that most of the pieces of the model are standard techniques. Overall, the paper does not achieve the minimum score required to be published at NIPS.

~\\
\textbf{Reviewer \#1}

1. Please provide an "overall score" for this submission.

\textbf{6}: Marginally above the acceptance threshold. I tend to vote for accepting this submission, but rejecting it would not be that bad.

~\\
2. Please provide a "confidence score" for your assessment of this submission.

4: You are confident in your assessment, but not absolutely certain. It is unlikely, but not impossible, that you did not understand some parts of the submission or that you are unfamiliar with some pieces of related work.

~\\
3. Please provide detailed comments that explain your "overall score" and "confidence score" for this submission. You should summarize the main ideas of the submission and relate these ideas to previous work at NIPS and in other archival conferences and journals. You should then summarize the strengths and weaknesses of the submission, focusing on each of the following four criteria: quality, clarity, originality, and significance.

The paper describes a method for generating synthetic images of faces with different facial expressions. The authors propose a GAN-based architecture for the purpose, wherein, the facial identity and expression priors are disentangled and encoded in different latent spaces. The expression and identity facial latent vectors are concatenated and reconstructed by the generator to preserve the identity and change the facial expression. The facial expressions are represented by 68 facial fiducial points and encoded into a latent space. The authors use reconstruction losses for the input facial fiducial points and facial images along with adversarial training for the entire algorithm. Overall, neither the problem for synthesizing facial expressions, nor its proposed solution using GANs is novel. The paper is also specific to facial analysis and hence not directly applicable to many different problem domains. Hence the paper has decent novelty and some interesting findings, but is not earth-shattering or likely to be of broad impact.

\qquad In contrast to the prior work, the claimed superiority of this method lies in the fact that it allows to learn a Lipschitz continuous semantic latent space of facial expressions instead of a discrete one. The latter is what previous works have learnt via one-hot encoding of the facial expression classes. By comparing their approach to that of [23], the authors show empirically the value of learning a latent embedding of the facial fiducial points for continuous transitions of expressions versus using the facial fiducial points directly as inputs along with the facial image for the GAN-based approach. The latter they argue is not guaranteed to be Lipschitz continuous. This is an interesting finding.

\qquad One other finding of this work is that using the contrastive loss for encoding facial expressions helps to learn a more semantically meaningful expression latent space and hence helps to produce better expression synthesis results on moving linearly on the expression manifold. However, the authors’ use of the contrastive loss is at odds with their goal of trying to learn a *continuous* latent expression space. The contrastive loss tends to create discontinuous clusters for discrete facial expressions classes, which is exactly what the authors observe in the figure 4(b) on visualizing their learnt latent space. This is also something that the authors set out to avoid in the first place. How then do the authors explain the helpfulness of the constructive loss in learning a continuous latent expression space? This contradictory observation needs to explored further and clearly addressed and explained in the paper. I would have imagined that something like a variational auto encoder with a KL divergence loss term would have helped to create a more continuously varying latent expression space.

\qquad Other than that, the paper is well written, includes sufficient experimental evidence to demonstrate the superior performance of their method versus previous work, and includes sufficient details in the paper and the supplementary document of the network and training for someone to be able to replicate it.

~\\
4. How confident are you that this submission could be reproduced by others, assuming equal access to data and resources?

3: Very confident
 
~\\
\textbf{Reviewer \#2}

1. Please provide an "overall score" for this submission.

\textbf{6}: Marginally above the acceptance threshold. I tend to vote for accepting this submission, but rejecting it would not be that bad.

~\\
2. Please provide a "confidence score" for your assessment of this submission.

4: You are confident in your assessment, but not absolutely certain. It is unlikely, but not impossible, that you did not understand some parts of the submission or that you are unfamiliar with some pieces of related work.

~\\
3. Please provide detailed comments that explain your "overall score" and "confidence score" for this submission. You should summarize the main ideas of the submission and relate these ideas to previous work at NIPS and in other archival conferences and journals. You should then summarize the strengths and weaknesses of the submission, focusing on each of the following four criteria: quality, clarity, originality, and significance.

The paper introduces Geometry-Contrastive Generative Adversarial Network (GC-GAN), a novel deep architecture for transferring facial emotions across different subjects. The proposed architecture is based on the idea of decoupling the identity information with the motion information in the latent representation of the main network. A novel contrastive loss is employed to learn the latent motion information from landmark images. The proposed approach is evaluated on three common face image benchmarks.

Strengths:

- This paper, together with [23, 28], is one of the first attempts to inject geometry information into GAN models for facial expression transfer. This is an interesting and challenging problem.

- The idea of using contrastive learning to compute the latent representation from the landmark images is new, despite contrastive learning have been exploited in many other applications

- The experimental evaluation consider three different benchmarks where face have different characteristics. The proposed approach is shown to be more effective than previous method [23] on the proposed task.

- The paper is well written and the proposed approach clearly explained.

Weaknesses:

- In my opinion the experimental evaluation can be improved. Specifically, when analyzing the dynamics of face sequences in section 4.3 a comparison with [28] can be added. This will confirm the effectiveness of the proposed contrastive loss.

- In Table 2 it is not clear what SPG is. Which method was used exactly?

- The results from a user study could be added, as SSIM may be not the best metric for analyzing generated faces.

~\\
4. How confident are you that this submission could be reproduced by others, assuming equal access to data and resources?

2: Somewhat confident

~\\
\textbf{Reviewer \#4}

1. Please provide an "overall score" for this submission.

\textbf{4}: An okay submission, but not good enough; a reject. I vote for rejecting this submission, although I would not be upset if it were accepted.

2. Please provide a "confidence score" for your assessment of this submission.

4: You are confident in your assessment, but not absolutely certain. It is unlikely, but not impossible, that you did not understand some parts of the submission or that you are unfamiliar with some pieces of related work.

3. Please provide detailed comments that explain your "overall score" and "confidence score" for this submission. You should summarize the main ideas of the submission and relate these ideas to previous work at NIPS and in other archival conferences and journals. You should then summarize the strengths and weaknesses of the submission, focusing on each of the following four criteria: quality, clarity, originality, and significance.

This paper presented an framework - GC-GAN for facial expression transfer/generation. The proposed method consists of a facial geometry embedding network and a conditional GAN networks. The experimental on several datasets show the effectiveness of GC-GAN.

Some strong points of the paper:

1. the integration of the facial geometry embedding network is somehow novel

2. the presentation of the paper is good

Some weak points of the paper:

1.the novelty contribution of the overall framework is minor. The generator/discriminator just follow general conditional GANs. The geometry-guided idea have been explored a lot in previous works (such as G2GAN,GAGAN).

2. the experimental part is not strong: a) the model has many hyper-parameters/pretrianed modules setting, making it difficult to reproduce the results. The details of those are not described clearly; b) just few baselines are compared in the results. There should be more about GANs with structured conditions.

3. The analysis of geometry-contrastive learning, i.e., the facial embedding network is not convincing. Section 4.3 can not give a convincing quantitative result. The measure in Section 4.4 only takes the overall measure but fails to show how the generated samples have better representation regarding to the continuity.

Some related and important references are missing about GANs:

a. Disentangled Representation Learning GAN for Pose-Invariant Face Recognition

b. Disentangled Person Image Generation

c. Pose Guided Person Image Generation

4. How confident are you that this submission could be reproduced by others, assuming equal access to data and resources?

2: Somewhat confident

\section{Rebuttal}
\textbf{Response to Reviewer \#1}

\textbf{Q}: About the novelty and potential application to other problem domains.

\textit{\textbf{A}: Yes, the problem for synthesizing facial expressions is not new, but how to integrate continuity of facial expressions in GANs hasn’t been mentioned or studied carefully before, thus the detailed problem and our solution are novel to a certain extent. Facial expressions are intrinsically continuous but always annotated with discrete labels, so our method can also be applied in similar problems such as the pose of objects, the length of hair, the height of heels and so on.}

\textbf{Q}: About the effect of contrastive learning for semantically continuous representation.

\textit{\textbf{A}: Contrastive learning has been widely used in face identification such as FaceNet, VGG-Face and so on. Face identification is a typical classification problem, however, these networks trained for it also generalize well on face verification where faces are unseen. Similarly, such learning mechanism enables the latent space to accommodate unseen emotion states. As seen in Figure 4(b), the data scatters of disgust and squint are not separated well with each other due to the intrinsic geometry similarity although they are different expressions. In Figure 4(c), we present an embedding line corresponding to the transition from smile to scream of the same person obtained by equal interval sampling on facial landmarks. It can be observed that the eyes are closing and the mouth is opening, which is exactly the embodiment of continuity. We try to give a general architecture of continuous latent space in cGANs and have embedded an autoencoder in current work, and we will test KL divergence loss term in the future. Thanks!}

~\\

\textbf{Response to Reviewer \#2}

\textbf{Q}: The comparison with [28].

\textit{\textbf{A}: [28] proposes a method to generate facial image sequences of diverse smiles, which is an excellent work. We’ll add this comparison in terms of analyzing the dynamics of face sequences in our future work.}

\textbf{Q}: The explanation of SPG and user study.

\textit{\textbf{A}: SPG is the abbreviation of “similar to PG$^2$ and G2GAN”, which means the method of concatenating geometry and images directly as in both PG$^2$ and G2GAN. PG$^2$ mainly focuses on the generation of persons of different poses and the input facial images of G2GAN should be neutral expressions. Besides, the ways of training GANs, neural network architectures and loss functions among PG$^2$, G2GAN and GC-GAN are different. It’s improper to directly compare their methods with ours considering the reasons mentioned above, so we construct such configuration for fair comparison. We also took user study into consideration. However, user study is somewhat subjective and may cause some trouble if other people want to compare their methods with ours. We’ll conduct a user study if necessary. Thanks for your advice!}

~\\

\textbf{Response to Reviewer \#4}

\textbf{Q}: The novelty contribution of the overall framework is minor.

\textit{\textbf{A}: Instead of following the general conditional GAN, we concentrate on how to establish semantically continuous conditions into GANs, and formal description of Lipschitz continuity was given in Section 3.2. Despite geometry-guided ideas have been explored directly in previous works, we are the first one to embed geometry information in contrastive learning to maintain continuity into cGANs.}

\textbf{Q}: The experimental part is not strong.

\textit{\textbf{A}: 
	a) The detailed architecture and the key parameters are provided in supplementary materials. This work can be easily reproduced and code will be available upon publication.}
\textit{b) Two types of baselines have been compared. First, a standard one-hot code based cGAN CDAAE [34] and also an ablation study have been conducted in Section 4.2. Second, we compare our method with two state-of-the-art methods PG$^2$ [18] and G2GAN [23] in Section 4.4. Three different benchmarks have been tested both qualitatively and quantitatively.}

\textbf{Q}: The analysis of geometry-contrastive learning.

\textit{\textbf{A}: Firstly, we evaluate the Lipschitz continuity of generated faces over facial expression conditions in Section 4.3, and the quantitative result is given in Figure 3. Besides, the embedded manifold is visualized qualitatively in Figure 4 for intuitive interpretation. Secondly, in Section 4.4, we mainly demonstrate how geometry-contrastive learning handles the misalignments between different persons and the generalization on the spontaneous emotion states.}

\textbf{Q}: Some related and important references are missing about GANs.

\textit{\textbf{A}:  Paper (c) has been cited as [18] in our paper. Paper (b) hasn’t been officially published before the submission deadline and doesn’t get much improvements compared with paper (c). Paper (a) concentrates on face frontalization and pose-invariant face recognition, which doesn’t have close relationship with our work.}

\end{document}